\documentclass[conference]{IEEEtran}
\IEEEoverridecommandlockouts
\usepackage{cite}
\usepackage{amsmath,amssymb,amsfonts}
\usepackage{algorithmic}
\usepackage{graphicx}
\usepackage{textcomp}
\usepackage{xcolor}
\def\BibTeX{{\rm B\kern-.05em{\sc i\kern-.025em b}\kern-.08em
    T\kern-.1667em\lower.7ex\hbox{E}\kern-.125emX}}

\begin{document}

\title{Corpus Characterization and Inverse Constitutional Fine-Tuning for Style-Aware Radiology Reports}

\author{\IEEEauthorblockN{Sarah Li}
\IEEEauthorblockA{\textit{McLean High School} \\
sarah.yingqi.li@gmail.com}
\and
\IEEEauthorblockN{Elijah Renner}
\IEEEauthorblockA{\textit{Stanford University}\\
elijahcrenner@gmail.com}
\and
\IEEEauthorblockN{Rayan Ansari}
\IEEEauthorblockA{\textit{Stanford Cardiovascular Institute} \\
\textit{Stanford University}\\
rayansri@stanford.edu}
\and
\IEEEauthorblockN{Alaa Youssef}
\IEEEauthorblockA{\textit{Department of Radiology} \\
\textit{Stanford School of Medicine)}\\
ayoussef@stanford.edu}
}

\maketitle

\begin{abstract}
Automated radiology report generation has advanced rapidly in diagnostic accuracy, yet generated reports frequently diverge from the stylistic conventions of authentic radiologist writing in structure, diction, and uncertainty language, a gap which has direct implications for clinician trust and user experience. To address this, we characterize stylistic variation across 2,000 reports from the CheXpert Plus dataset using Bio-ClinicalBERT embeddings, UMAP dimensionality reduction, and HDBSCAN clustering, identifying five distinct reporting patterns differing in pathology focus, narrative structure, and lexical preference. Drawing on these findings, we adapt the inverse constitutional AI framework to derive a style-focused constitution from radiologist-written report pairs without requiring a formal preference dataset. This constitution, encoding conventions of tone, diction, uncertainty calibration, and report structure, is incorporated into the supervised fine-tuning of a MedGemma-4B base model on 25,245 CheXpert Plus training pairs. Constitutional fine-tuning produces a substantial increases in text alignment (BLEU-4: 0.006→0.308; ROUGE-L: 0.171→0.484) relative to the untuned baseline. These gains show a qualitative shift in structural and lexical alignment rather than marginal improvement, as the baseline model produces near-zero scores due to format mismatch. Overall, we establish corpus-level style characterization and constitutional modeling as an effective and data-efficient strategy for producing radiology reports that conform to authentic radiologist writing conventions.
\end{abstract}


\section{Introduction}

\subsection{Problem Statement}


Radiology reports are the primary medium through which radiologists translate imaging findings into clinical guidance, synthesizing observations, interpretations, and diagnostic impressions into a document that drives downstream care decisions ~\cite{pahadia_radiology_2020}. Effective reporting demands not only diagnostic accuracy, but also proper grammar, active voice, definitive phrasing, and logical organization to ensure findings are correctly understood and acted upon ~\cite{pahadia_radiology_2020}. The clinical stakes of stylistic failure are substantial, as failure to communicate results clearly and effectively is among the three most common grounds for malpractice suits against radiologists~\cite{wilcox_written_2006}. Still, style remains a largely underexplored dimension of automated report generation.

Recent advances in large language models (LLMs) and vision-language models have enabled automated systems to generate chest radiograph reports with increasingly strong diagnostic content \cite{van_veen_adapted_2024, zhang_comparison_2023}. In prospective clinical settings, AI-assisted draft reporting has been shown to reduce documentation time by over 15\% while maintaining clinical
quality \cite{huang_efficiency_2025}, and AI-generated reports have been rated as equivalent or superior to human-written reports in accuracy and radiologist confidence \cite{rajmohamed_evaluating_2025}. However, the field has largely focused on clinical accuracy by measuring CheXbert label agreement and RadGraph F1 while paying comparatively little attention to whether generated reports conform to the stylistic conventions of authentic radiologist writing. This gap is clinically meaningful: a report that is factually complete but stylistically misaligned (e.g. using passive constructions, excessive hedging, non-standard section ordering, or uncharacteristic vocabulary) may impede clinical decision-making and undermine practitioner trust in automated outputs. At the same time, standard lexical overlap metrics such as BLEU and ROUGE penalize semantically valid reports that differ in surface form without assessing clinical correctness~\cite{ostmeier_green_2024, li_reevalmed_2025}. A method that explicitly characterizes and enforces writing style is therefore needed to bridge this modeling and evaluation gap.

\subsection{Related Works}

\textbf{NLP for radiology reports.}
A substantial body of work applies NLP to radiology reports for tasks including label extraction, finding classification, and report summarization~\cite{casey_systematic_2021}. Domain-adapted encoders such as Bio-ClinicalBERT~\cite{alsentzer_publicly_2019} and RadBERT~\cite{yan_radbert_2022} improved performance over general-domain BERT models on radiology-specific tasks including report summarization and abnormal sentence classification. Structured labeling tools such as CheXbert~\cite{smit_chexbert_2020} enable automated extraction of 14 radiographic findings from free-text reports. NLP pipelines have been applied to a wide range of radiology tasks, including detecting discrepancies between preliminary and final reports~\cite{wang_automated_2026}, characterizing the change and clinical significance of findings over time~\cite{hassanpour_characterization_2017}, and identifying incidental findings such as pulmonary nodules~\cite{grolleau_incidental_2024}. Template-based generation approaches, such as Replace and Report~\cite{kale_replace_2023}, construct reports by identifying abnormal findings and substituting them into normal report templates, achieving substantial improvements on BLEU, ROUGE-L, METEOR, and CIDEr over end-to-end baselines on the IU X-Ray and MIMIC-CXR datasets.

\textbf{Style-aware generation.}
Despite the clinical importance of writing conventions, style has received comparatively little attention in automated report generation. Yan et al.~\cite{yan_style-aware_2023} propose a two-step approach that disentangles content from style: a dedicated model extracts clinical entities as a serialized RadGraph representation, and a frozen LLM uses few-shot radiologist examples to verbalize this content in the target writer's style. Their human evaluation found that AI-generated reports were indistinguishable from authentic reports in over 75\% of cases, though standard NLP metrics did not improve. More recently, Delbrouck et al.~\cite{delbrouck_automated_2025} introduced Structured Radiology Report Generation (SRRG), which uses GPT-4 to reformat free-text reports from MIMIC-CXR and CheXpert Plus~\cite{chambon_chexpert_2024} into a standardized anatomical section structure, arguing that stylistic variability in existing datasets undermines both generation consistency and evaluation reliability.

\textbf{Constitutional and alignment-based methods.}
Constitutional AI (CAI) \cite{bai_constitutional_2022} provides a principled framework for embedding behavioral guidelines into language models via a written set of natural-language principles. Findeis et al.~\cite{findeis_inverse_2025} introduced inverse constitutional AI (ICAI), which reverses this process: given a corpus of pairwise preference data, ICAI uses an LLM to generate candidate principles, clusters them by semantic similarity to eliminate redundancy, and validates each principle's predictive power on the data. Henneking and Beger ~\cite{henneking_decoding_2025} further refined the ICAI algorithm, demonstrating that improved principle generation and embedding processes enhance the accuracy and generalizability of extracted constitutions across synthetic and real-world datasets. Our work adapts this framework to the medical domain, extracting stylistic principles from radiologist-written reports without requiring a formal preference dataset.

\textbf{Evaluation metrics.}
Prior work has noted fundamental limitations of BLEU and ROUGE for radiology report evaluation~\cite{ostmeier_green_2024, yu_evaluating_2023}: lexical overlap metrics cannot distinguish stylistically appropriate paraphrases from clinically erroneous content and do not correlate reliably with expert radiologist preferences. ReEvalMed~\cite{li_reevalmed_2025} further demonstrates that generated reports frequently score well on standard metrics while failing to capture clinically significant errors or distinguish error severity levels. Clinical metrics such as RadGraph F1~\cite{jain_radgraph_2021} and CheXbert label agreement~\cite{smit_chexbert_2020} address factual correctness but do not capture style. The GREEN metric~\cite{ostmeier_green_2024}, which uses LLMs to identify and categorize clinically significant errors, offers a more interpretable and expert-aligned evaluation. Benchmarks such as CXPMRG-Bench~\cite{wang_cxpmrg-bench_2024} provide standardized baselines for comparing generation models on CheXpert Plus across these metrics. Our work uses BLEU-4 and ROUGE-L as proxies for style alignment, and we discuss the implications of this choice in the context of these known limitations.

\subsection{Contributions}

This work makes the following contributions:
\begin{itemize}
    \item We present a systematic characterization of stylistic variation across a subset of CheXpert Plus radiology reports~\cite{chambon_chexpert_2024}, identifying five distinct reporting clusters via Bio-ClinicalBERT embeddings~\cite{alsentzer_publicly_2019}, UMAP dimensionality reduction~\cite{mcinnes_umap_2018}, and HDBSCAN clustering~\cite{mcinnes_hdbscan_2017}.

    \item We construct a style-focused constitution derived using principles of inverse constitutional AI~\cite{findeis_inverse_2025, henneking_decoding_2025}, capturing conventions of tone, diction, uncertainty expression, structure, and clinical ethics without requiring a formal preference dataset.

    \item We demonstrate that constitutional supervised fine-tuning of MedGemma-4B~\cite{sellergren_medgemma_2026} yields substantial improvements in lexical alignment with radiologist writing, achieving a 5{,}000\% increase in BLEU-4 and a 183\% increase in ROUGE-L relative to the untuned baseline.

    \item We provide a replicable methodology for extracting domain-specific style constitutions from paired clinical text corpora, with potential applicability beyond radiology to other structured documentation tasks.
\end{itemize}

\section{Methods}

\subsection{Data Description}
We use the CheXpert Plus dataset from the Stanford AIMI Center \cite{chambon_chexpert_2024}. CheXpert Plus contains 223,462 chest X-rays paired with de-identified, full-text radiology reports, providing image-report pairs, of which the reports were essential to style analysis.

Using CheXbert \cite{smit_chexbert_2020} to label radiologist reports, we analyze the distribution of labels. The labels correspond to positive, negative, and uncertain mentions of each radiographic finding.

\begin{table}[htbp]
\caption{CheXbert Label Distribution Across Radiologist Reports}
\label{tab:chexbert-label-distribution}
\centering
\scriptsize
\begin{tabular}{lrrr}
\hline
Finding & Positive & Negative & Uncertain \\
\hline
Enlarged Cardiomediastinum & 7559 & 22522 & 15021 \\
Cardiomegaly & 30558 & 16160 & 3921 \\
Lung Opacity & 102950 & 5081 & 330 \\
Lung Lesion & 9368 & 1349 & 1557 \\
Edema & 53011 & 21229 & 12217 \\
Consolidation & 13702 & 31452 & 26627 \\
Pneumonia & 4847 & 3322 & 19367 \\
Atelectasis & 33851 & 727 & 34401 \\
Pneumothorax & 17879 & 58631 & 2418 \\
Pleural Effusion & 89267 & 36290 & 7614 \\
Pleural Other & 3957 & 134 & 2724 \\
Fracture & 8706 & 3927 & 494 \\
Support Devices & 115892 & 3978 & 54 \\
No Finding & 21259 & 0 & 0 \\
\hline
\end{tabular}
\end{table}

\subsection{Report Characterization}
To characterize the stylistic and disease-level variation present in the CheXpert Plus reports, we used CheXbert ~\cite{smit_chexbert_2020} to label a representative sample of radiologist-written reports, producing positive, negative, and uncertain annotations for each of 14 radiographic findings. We then generated dense semantic embeddings for a random subset of 2,000 reports using Bio-ClinicalBERT~\cite{alsentzer_publicly_2019}, a domain-adapted encoder pre-trained on clinical discharge summaries. To make the high-dimensional embedding space easier to cluster, we applied UMAP~\cite{mcinnes_umap_2018} for nonlinear dimensionality reduction, followed by HDBSCAN~\cite{mcinnes_hdbscan_2017}, a density-based algorithm that identifies clusters of arbitrary shape without requiring a predefined cluster count. Five discrete clusters emerged from this procedure. For each cluster, we examined the distribution of CheXbert labels, the distribution of specific radiographic findings, and patterns in lexical features, creating an empirical basis for the stylistic components later encoded in the constitution.

\begin{figure}[htbp]
\centerline{\includegraphics[width=\linewidth]{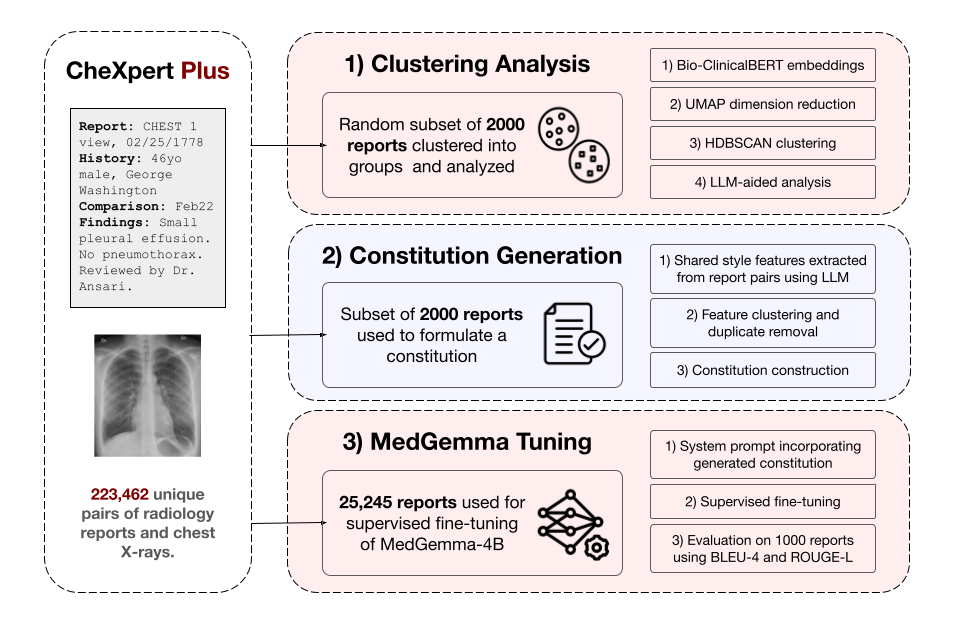}}%
\caption{Pipeline for the three main methods: report characterization, constitution generation, and MedGemma fine-tuning.}
\label{fig:framework}
\end{figure}

\subsection{Constitution Generation}
We leveraged the principles of inverse constitutional AI to create a constitution governing the style of generated reports. Because we lacked a preference dataset, we adapted the framework of Findeis et al \cite{findeis_inverse_2025} by substituting pairs of radiologist-written reports for conventional preference annotations. An LLM was prompted to extract shared stylistic features from each report pair, including recurring syntactic patterns, domain-specific vocabulary, phrasing, and structure. After parsing all extracted features, we clustered similar features together and removed duplicates, following the generate-cluster-deduplicate pipeline refined by Henneking and Beger~\cite{henneking_decoding_2025}, to produce a compact and non-redundant set of guidelines. Ultimately, the constitution incorporated instructions for output and writing guidelines, as well as incorporating ethical considerations such as upholding confidentiality and refraining from overconfident speculation.

\subsection{MedGemma Supervised Fine-Tuning}
We performed supervised fine-tuning (SFT) of MedGemma-4B~\cite{sellergren_medgemma_2026} on a subset of 25,245 image-report training pairs drawn from CheXpert Plus~\cite{chambon_chexpert_2024}. A static system prompt, shared across all training examples, embedded the generated constitution along with explicit structural directives instructing the model to produce reports in five standardized sections: \texttt{[NARRATIVE]}, \texttt{[HISTORY]}, \texttt{[COMPARISON]}, \texttt{[IMPRESSION]}, and \texttt{[SUMMARY]}. This structured output format facilitates section-level comparison against reference reports during evaluation and aligns with calls in the literature for more consistent, structured clinical reporting~\cite{delbrouck_automated_2025}.

\subsection{Evaluation}
To quantify the effect of constitutional fine-tuning on output quality, we evaluated 2,000 generated reports for unseen chest X-ray images against held-out radiologist-written references using BLEU-4~\cite{papineni_etal_2002_bleu} and ROUGE-L~\cite{lin_2004_rouge} as measures of lexical alignment. BLEU-4 captures four-gram precision between generated and reference text, making it sensitive to structural and terminological conformity, while ROUGE-L measures the longest common subsequence, reflecting sentence-level fluency and narrative flow. Both metrics were computed on outputs from the base MedGemma-4B model and the constitutionally fine-tuned model, with the difference between these scores serving as the primary measure of style alignment improvement.

\section{Results}

\subsection{Characterization of Radiology Reports}

HDBSCAN clustering of UMAP-reduced report embeddings yielded five discrete groups with distinct pathological and stylistic profiles. Cluster $-1$, was enriched for acute and inflammatory conditions, showing the highest z-scores for Pneumonia ($+1.60$) and Consolidation ($+1.59$). Cluster 0 was characterized by stable reporting, with enrichment for Atelectasis ($+1.28$) and Pleural Other ($+1.38$) and language describing unchanged findings. Cluster 1, the largest group by far, showed the broadest pathological distribution, with enrichment for Cardiomegaly ($+1.25$), Fracture ($+1.26$), and Support Devices ($+1.03$), showing the heterogeneous case mix typical of general chest radiography. Cluster 2 diverged from the others stylistically as much as pathologically: it was the least concise, and reports had more extended discussion of clinical implications instead of direct observation. Cluster 3 was the most pathologically specific, with strong enrichment for Pneumothorax ($+1.61$) and Pleural Effusion ($+1.31$), and reports that consistently focused on medical devices. Conciseness and direct observational language were shared across all clusters except Cluster 2, which informed several principles of the style constitution.

\begin{table}[htbp]
\caption{Distributions of CheXpert-extracted Clusters}
\label{tab:cluster-distribution}
\centering
\begin{tabular}{lrrr}
\hline
Cluster & Count \\
\hline
-1 & 32 \\
0 & 117 \\
1 & 1529 \\
2 & 107 \\
3 & 215 \\
\hline
\end{tabular}
\end{table}

\begin{figure}[htbp]
\centerline{\includegraphics[width=\linewidth]{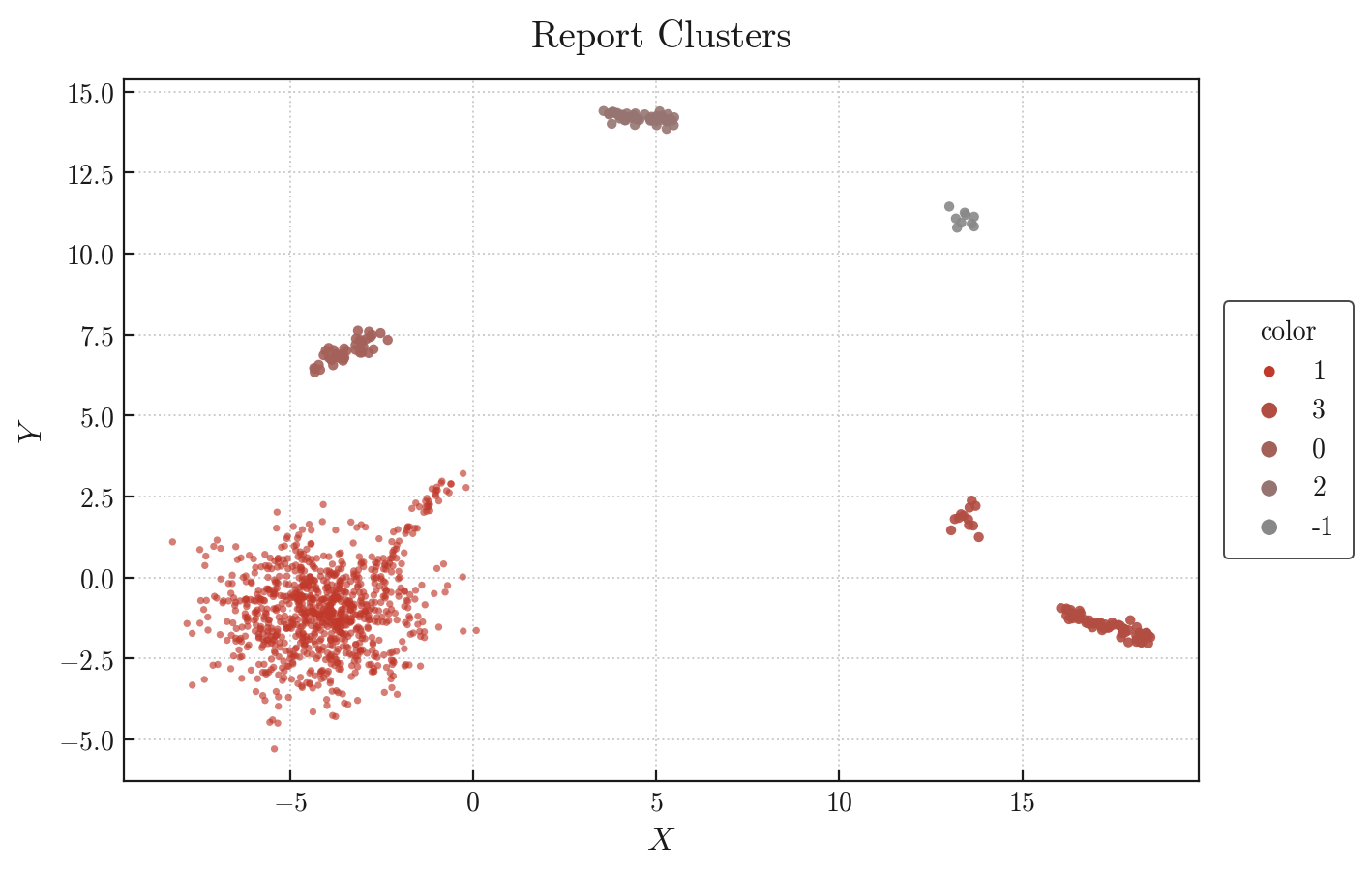}}%
\caption{HBDSCAN clustering on the UMAP-reduced embeddings of a subset of 2000 CheXpert reports. 5 clusters were identified.}
\label{fig:report-clusters}
\end{figure}

\begin{figure}[htbp]
\centerline{\includegraphics[width=\linewidth]{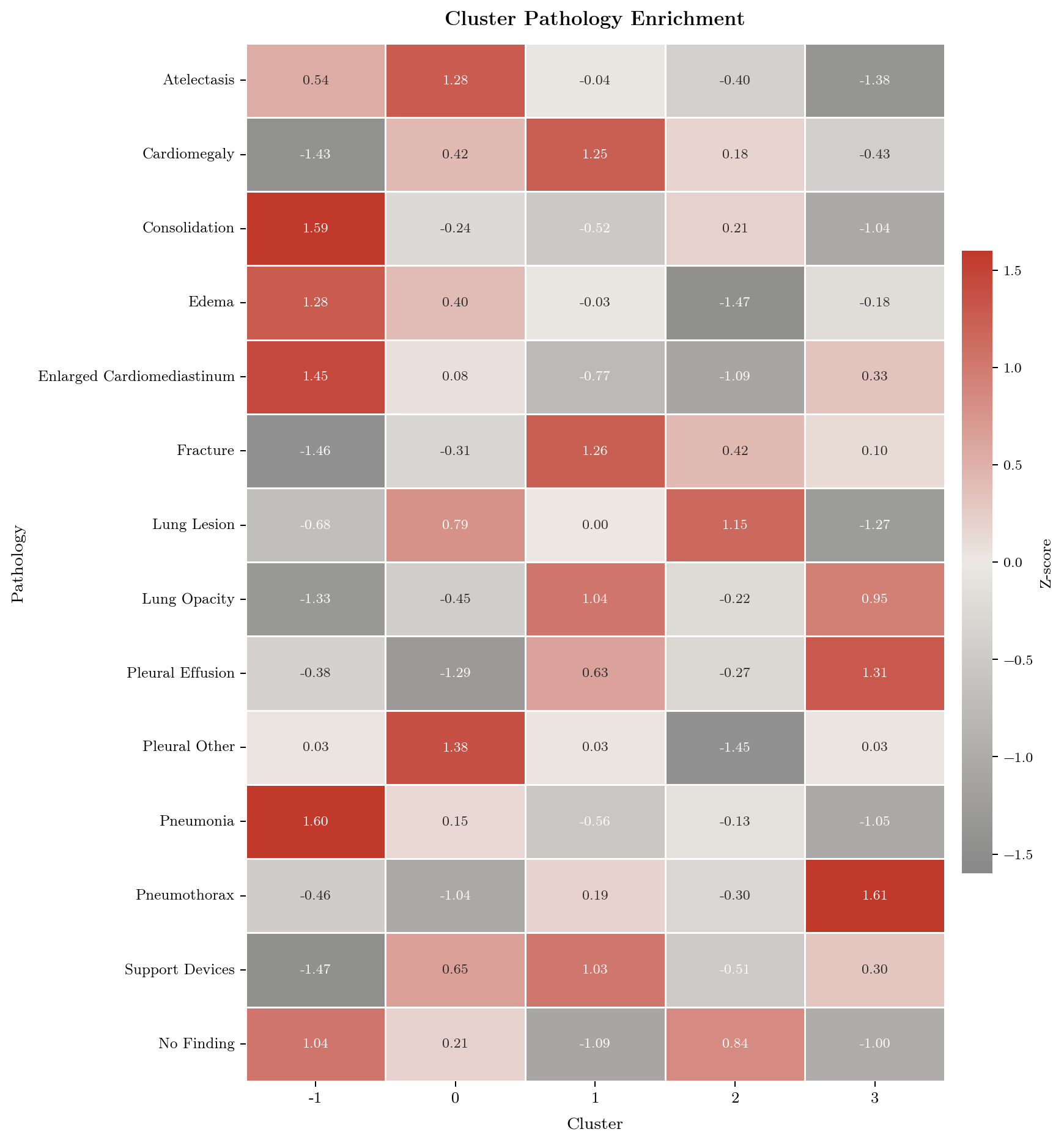}}%
\caption{Pathology enrichment for the 5 HDBSCAN-identified clusters.}
\label{fig:report-clusters-enrichment}
\end{figure}

\subsection{Inverse Generation of AI Constitution}

The generate-cluster-deduplicate pipeline yielded principles organizing around three recurring dimensions: structural conventions, lexical and syntactic preferences, and conciseness. Structural conventions were the most consistent cross-cluster finding. Radiologist-written reports follow predictable section ordering, use standardized headers, and separate the description of findings from their interpretation. Lexical and syntactic patterns were also prominent, including preferences for active voice, specific anatomical measurement conventions, and characteristic clinical terminology~\cite{pahadia_radiology_2020}. Conciseness, which entailed minimizing redundant language while preserving diagnostic completeness, characterized the majority of clusters.

The resulting constitution was incorporated into the SFT system prompt alongside structural output instructions specifying five required report sections: \texttt{[NARRATIVE]}, \texttt{[HISTORY]}, \texttt{[COMPARISON]}, \texttt{[IMPRESSION]}, and \texttt{[SUMMARY]}. Ethical guidelines were also embedded: the model is instructed to protect patient confidentiality and to avoid unsupported clinical speculation, practices aligned with established standards for responsible radiology reporting~\cite{wilcox_written_2006}.

This approach differs from prior style-injection methods~\cite{yan_style-aware_2023} in that it encodes aggregate corpus-level conventions rather than any individual radiologist's idiosyncrasies, making it more suitable for general-purpose deployment. It also differs from the structured reformatting approach of Delbrouck et al.~\cite{delbrouck_automated_2025} in that style principles are derived inductively from the data rather than imposed through a top-down GPT-4 reformatting pass, preserving the natural linguistic variation of authentic radiologist writing while still enforcing core conventions.

\subsection{Report Evaluation}

Constitutional SFT of MedGemma-4B~\cite{sellergren_medgemma_2026} yielded substantial improvements in lexical alignment with radiologist-written reference reports. BLEU-4 increased by 5{,}000\% from 0.006 to 0.308 and ROUGE-L increased by 183\% from 0.171 to 0.484 relative to the untuned baseline.

\begin{figure}[htbp]
\centerline{\includegraphics[width=\linewidth]{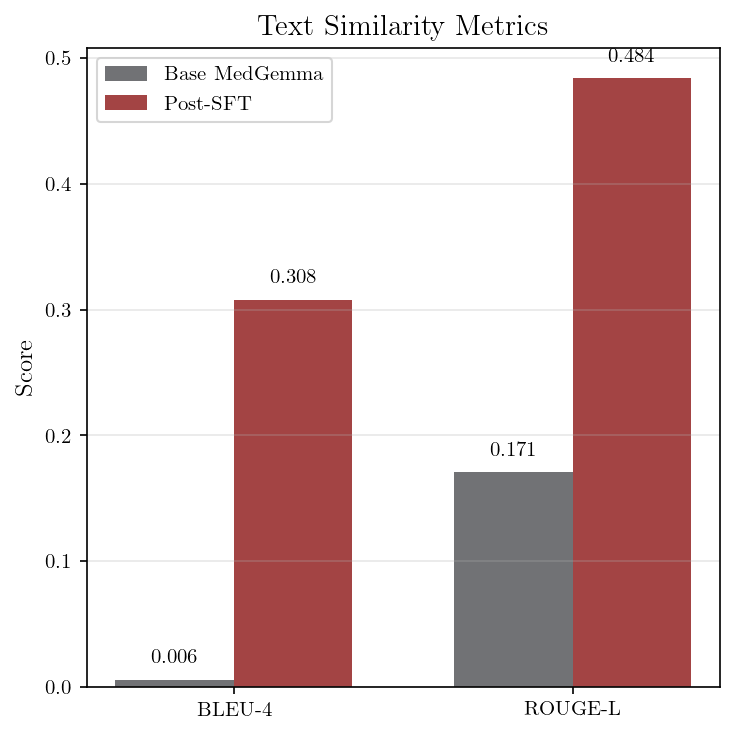}}%
\caption{Comparison of BLEU-4 and ROUGE-L between the base MedGemma model and the fine-tuned model.}
\label{fig:text_similarity_metrics}
\end{figure}

\begin{figure}[htbp]
\centerline{\includegraphics[width=\linewidth]{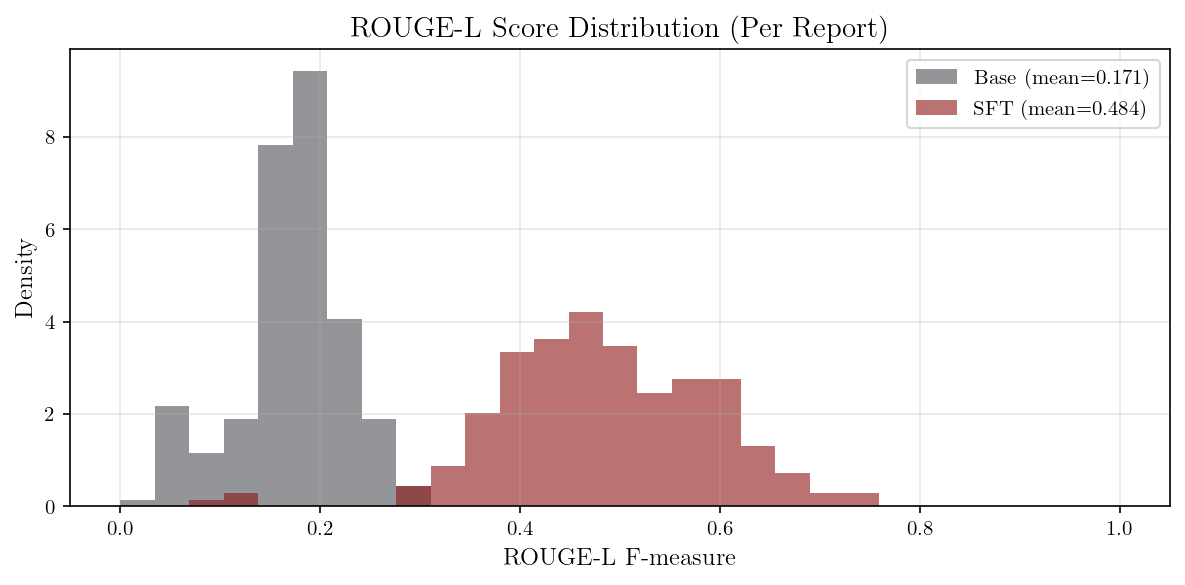}}%
\caption{Distribution of the ROUGE-L score per report, comparing the base MedGemma model and the fine-tuned model.}
\label{fig:rouge_dist}
\end{figure}

The magnitude of the BLEU-4 improvement reflects a qualitative shift in output format rather than marginal gains in lexical precision. The untuned MedGemma model, while medically capable, generates outputs that do not conform to the five-section structure of CheXpert Plus reference reports or to the specific lexical patterns of the training corpus. BLEU-4, which requires exact four-gram matches between generated and reference text, is highly sensitive to structural and vocabulary differences. Even a diagnostically accurate report expressed in different section headers or phrasing will score near zero. Constitutional SFT resolves this by training the model to reproduce the expected report framework, terminology, and stylistic register, producing a large increase in four-gram precision. The more moderate ROUGE-L improvement reflects gains at the sentence level via longest-common-subsequence overlap, consistent with a model that has acquired the structural flow of authentic reports while still producing some paraphrastic rather than verbatim phrasing.

These results should be interpreted as evidence of style alignment rather than clinical accuracy improvement. As noted by Ostmeier et al.~\cite{ostmeier_green_2024}, Yu et al.~\cite{yu_evaluating_2023}, and Li et al.~\cite{li_reevalmed_2025}, BLEU and ROUGE do not correlate reliably with radiologist preference or diagnostic correctness: a high-BLEU report may be stylistically authentic but clinically incomplete, while a low-BLEU report may contain all critical findings expressed in different words. Future evaluation using RadGraph F1~\cite{jain_radgraph_2021}, CheXbert label agreement~\cite{smit_chexbert_2020}, and the GREEN metric~\cite{ostmeier_green_2024} is necessary to assess whether constitutional fine-tuning preserves or improves clinical content alongside style.

It is instructive to compare our approach with Yan et al.~\cite{yan_style-aware_2023}, who use few-shot prompting to inject writing style without fine-tuning. Their method achieves strong human evaluation results---AI detection rates below 25\%---but does not improve standard NLP metrics. Our fine-tuning approach produces the complementary result: measurable, large-magnitude lexical alignment gains without yet having human evaluation data. These two methods are therefore not competing alternatives but complementary strategies: constitution-based SFT anchors the model to the distributional norms of the training corpus, while few-shot style prompting enables personalized adaptation at inference time. A combined approach may yield both broad stylistic consistency and individual writer fidelity.

\section{Conclusion}

We introduced a style-aware framework for radiology report generation grounded in systematic report characterization and constitutional modeling. Embedding-based clustering of a subset of CheXpert Plus reports~\cite{chambon_chexpert_2024} identified five distinct reporting styles differing in pathology focus, narrative structure, and lexical preferences. We constructed a constitution using principles of inverse constitutional AI~\cite{findeis_inverse_2025, henneking_decoding_2025} applied to radiologist-written report pairs, capturing conventions of tone, diction, uncertainty language, and report structure without requiring a formal preference dataset. Constitutional supervised fine-tuning of MedGemma-4B~\cite{sellergren_medgemma_2026} on 25{,}245 CheXpert Plus training pairs yielded a 5{,}000\% increase in BLEU-4 and a 183\% increase in ROUGE-L relative to the untuned baseline, demonstrating that explicitly modeling writing style substantially improves lexical alignment with authentic radiologist reports.

Several limitations warrant acknowledgment. Our evaluation is restricted to lexical overlap metrics, which measure stylistic surface alignment but not clinical accuracy; as established by Ostmeier et al.~\cite{ostmeier_green_2024}, Yu et al.~\cite{yu_evaluating_2023}, and Li et al.~\cite{li_reevalmed_2025}, such metrics can diverge substantially from expert clinical judgment. Validation using RadGraph F1~\cite{jain_radgraph_2021}, CheXbert label agreement~\cite{smit_chexbert_2020}, and the GREEN metric~\cite{ostmeier_green_2024} is necessary before drawing conclusions about clinical utility. Our characterization and constitution are also specific to the CheXpert Plus corpus and may not generalize to other institutions, imaging modalities, or report formats without recharacterization. Finally, human evaluation by practicing radiologists remains an important validation step to assess perceived authenticity and clinical appropriateness, as demonstrated by the strong human evaluation results achieved through few-shot style prompting in prior work~\cite{yan_style-aware_2023}.

Future work should address these gaps through clinical metric evaluation, radiologist reader studies, and extension to other imaging modalities and report types. We also intend to investigate whether constitutional fine-tuning can be combined with inference-time few-shot style prompting~\cite{yan_style-aware_2023} to achieve both general stylistic consistency and individual writer adaptation. More broadly, inducing a domain-specific style constitution from paired clinical text without requiring preference annotations may generalize to other structured clinical documentation tasks. The success of AI-assisted draft reporting in reducing documentation time while maintaining quality~\cite{huang_efficiency_2025, rajmohamed_evaluating_2025} underscores the clinical value of pursuing models that are not only diagnostically capable but also stylistically aligned with radiologist writing conventions.

\section*{Acknowledgment}
This work was part of a larger project that was organized through the Stanford AIMI Center. We acknowledge the Stanford AIMI Center for funding this research and providing the computer infrastructure to execute experiments. We also thank our mentors Dr. Alaa Youssef and Rayan Ansari for their invaluable directional feedback
throughout this research.


\bibliographystyle{unsrt}
\bibliography{refs}

\onecolumn
\newpage
\setlength{\parindent}{0pt}

\section{Appendix}
\subsection{System Prompt and Constitution}
You are an expert radiologist analyzing chest X-ray images. Generate a structured radiology report following the standard CheXpert format.

Use these exact section markers in your output:

[NARRATIVE]
Exam type and date.
[/NARRATIVE]

[HISTORY]
Patient demographics (age, sex) and clinical indication.
[/HISTORY]

[COMPARISON]
Reference to prior imaging if applicable. If none, state "No prior imaging available."
[/COMPARISON]

[IMPRESSION]
Numbered findings in order of clinical importance:
1. Critical/urgent findings (pneumothorax, new infiltrates)
2. Medical device positioning and status (tubes, lines, catheters)
3. Major parenchymal abnormalities (atelectasis, consolidation, masses)
4. Pleural findings (effusions, pneumothorax)
5. Cardiac findings (size, contour)
6. Minor findings and comparisons with prior studies
7. Normal or stable findings.
Include only findings relevant to the image. Do not pad with unnecessary items.
[/IMPRESSION]

[SUMMARY]
One of: NO SIGNIFICANT ABNORMALITY | ABNORMAL | CRITICAL
[/SUMMARY]


\bigskip
WRITING GUIDELINES:

- Use numbered lists in the IMPRESSION section
- Be specific with anatomical locations (e.g., "right lower lung zone", "left lung base")
- Include measurements when relevant
- Use standard radiology terminology
- Document negative findings when clinically relevant (e.g., "No evidence of pneumothorax")
- Use comparative language: "stable", "improved", "interval change", "unchanged"
- Use appropriate certainty qualifiers: "likely", "suggests", "possible" vs. definitive statements
- Maintain professional, objective tone
- Generate only the report content, no metadata (accession numbers, anonymization notes)
- Do not include any text outside the section markers

\bigskip
CONSTITUTION:

1. Style of Writing
- **Concise and Objective**: Reports are typically succinct, focusing on key findings without unnecessary elaboration. Each report presents information in a straightforward manner, often using bullet points or numbered lists.
- **Clinical and Technical**: The language is formal and technical, suitable for medical professionals, with a focus on clinical findings and implications.

2. Diction and Voice
- **Professional Tone**: The voice is authoritative, reflecting the expertise of the reporting physician. The use of medical terminology is prevalent, indicating a high level of familiarity with radiological concepts and anatomy.
- **Impersonal**: The reports maintain a third-person perspective, avoiding personal opinions or emotional language.

3. Sentence Lengths
- **Varied Lengths**: Sentences range from short, direct statements to longer, more complex sentences that may include multiple findings or conditions. However, brevity is preferred for clarity.
- **Use of Lists**: Many reports utilize lists to convey findings, which aids in readability and organization.
- Sentences are succinct and to the point, often dropping articles such as "a" or "the"

4. Common Sentence Structures
- **Standardized Findings Format**: Reports often follow a consistent structure, starting with findings followed by interpretations or recommendations. For example:
  - "1. [Finding]. 2. [Finding]."
  - "The [anatomical structure] is [condition]."
- **Descriptive Phrases**: Phrases such as "no significant change," "stable appearance," and "persistent findings" are frequently used to describe the status of various conditions.

5. Common Words or Phrases
- **Recurrent Terms**: Terms like "pneumothorax," "pleural effusion," "edema," "consolidation," "opacity," and "cardiomegaly" are commonly used across reports.
- **Descriptive Adjectives**: Adjectives like "mild," "moderate," "bilateral," "persistent," and "stable" often precede findings to indicate severity or stability.
- **Clinical Recommendations**: Phrases such as "recommend clinical correlation," "consider further evaluation," and "no new findings" are regularly included to guide subsequent actions.

6. Ethical Considerations
- **Confidentiality**: All reports must respect patient confidentiality and adhere to HIPAA regulations. Identifying patient information should be omitted or anonymized.
- **Clarity in Communication**: Reports should be clear and understandable to ensure that referring physicians can make informed decisions based on the findings.
- **Refusal to Speculate**: In cases of uncertainty or potential harm, the report should refrain from speculative conclusions, instead suggesting further investigation or clinical correlation.

7. Minimizing Harmfulness
- **Avoiding Alarmism**: Reports should avoid language that could unnecessarily alarm patients or referring doctors. Findings should be presented factually and without hyperbole.
- **Encouraging Follow-up**: Recommendations for follow-up imaging or clinical evaluation should be framed constructively, emphasizing the importance of monitoring changes rather than implying immediate concern.

\end{document}